\pdfoutput=1
\documentclass[11pt]{article}

\usepackage{graphicx}
\usepackage{booktabs}
\usepackage{tcolorbox}
\usepackage{xcolor}
\usepackage{subcaption}
\usepackage{pifont}
\usepackage{algorithm}
\usepackage{algpseudocode}
\usepackage{amsmath}
\usepackage{placeins}
\usepackage{caption}

\usepackage{ACL2023}

\usepackage{times}
\usepackage{latexsym}

\usepackage[T1]{fontenc}
\usepackage[utf8]{inputenc}

\usepackage{microtype}

\usepackage{inconsolata}

\title{Uncertainty Aware Learning for Language Model Alignment}

\author{Yikun Wang,\thanks{Yikun and Rui contributed equally.}\thanks{School of Computer Science, Fudan University, China.} Rui Zheng,\footnotemark[2]\footnotemark[1] Liang Ding,\thanks{The University of Sydney, Austrilia.}\thanks{~Corresponding author.} Qi Zhang,\footnotemark[2] Dahua Lin,\thanks{The Chinese University of Hong Kong, Hong Kong.}\thanks{Shanghai Artificial Intelligence Laboratory, China.} Dacheng Tao\thanks{Nanyang Technological University, Singapore.} \\[0.5em]
\texttt{\{yikunwang19, rzheng20, qz\}@fudan.edu.cn,}
\texttt{\{dhlin\}@ie.cuhk.edu.hk}\\
\texttt{\{liangding.liam, dacheng.tao\}@gmail.com}\\
}

\begin{document}
\maketitle
\begin{abstract}
% With the development of instruction-tuned large language models, effectively aligning pretrained foundation models has been a formidable task. Current alignment approaches often neglect the uncertainty property of data and treat all samples without discrimination, which causes poorer efficacy on benchmarks. In this paper, uncertainty serves as a core guidance in facilitating models to learn features for differential scenarios. We introduce the concept of leveraging uncertainty or information entropy elicited from capable large language models to align pretrained language models. Specifically, the uncertainty is mapped to the smooth values of label smoothing technique to sophisticatedly manipulate the loss penalty for each individual sample, which enhances alignment performance. Moreover, through in-depth analysis, uncertainty-aware learning is found to facilitate better clustering of token features in the feature space. Our uncertainty-aware approach outperforms supervised fine-tuning on popular benchmarks, exhibiting superior results and fewer instances of model degradation. The model trained in a mixed scenario has achieved improvements of 10.62\% on high-entropy tasks and 1.81\% on low-entropy tasks.
As instruction-tuned large language models (LLMs) evolve, aligning pretrained foundation models presents increasing challenges. Existing alignment strategies, which typically leverage diverse and high-quality data sources, often overlook the intrinsic uncertainty of tasks, learning all data samples equally. This may lead to suboptimal data efficiency and model performance. In response, we propose uncertainty-aware learning (UAL) to improve the model alignment of different task scenarios, by introducing the sample uncertainty (elicited from more capable LLMs). We implement UAL by a simple fashion -- adaptively setting the label smoothing value of training according to the uncertainty of individual samples. Analysis shows that our UAL indeed facilitates better token clustering in the feature space, validating our hypothesis. Extensive experiments on widely used benchmarks demonstrate that our UAL significantly and consistently outperforms standard supervised fine-tuning. Notably, LLMs aligned in a mixed scenario have achieved an average improvement of 10.62\% on high-entropy tasks (i.e., AlpacaEval leaderboard), and 1.81\% on complex low-entropy tasks (i.e., MetaMath and GSM8K).
\end{abstract}

\section{Introduction}

\begin{figure}[t]
\centering
\includegraphics[width=0.4\textwidth]{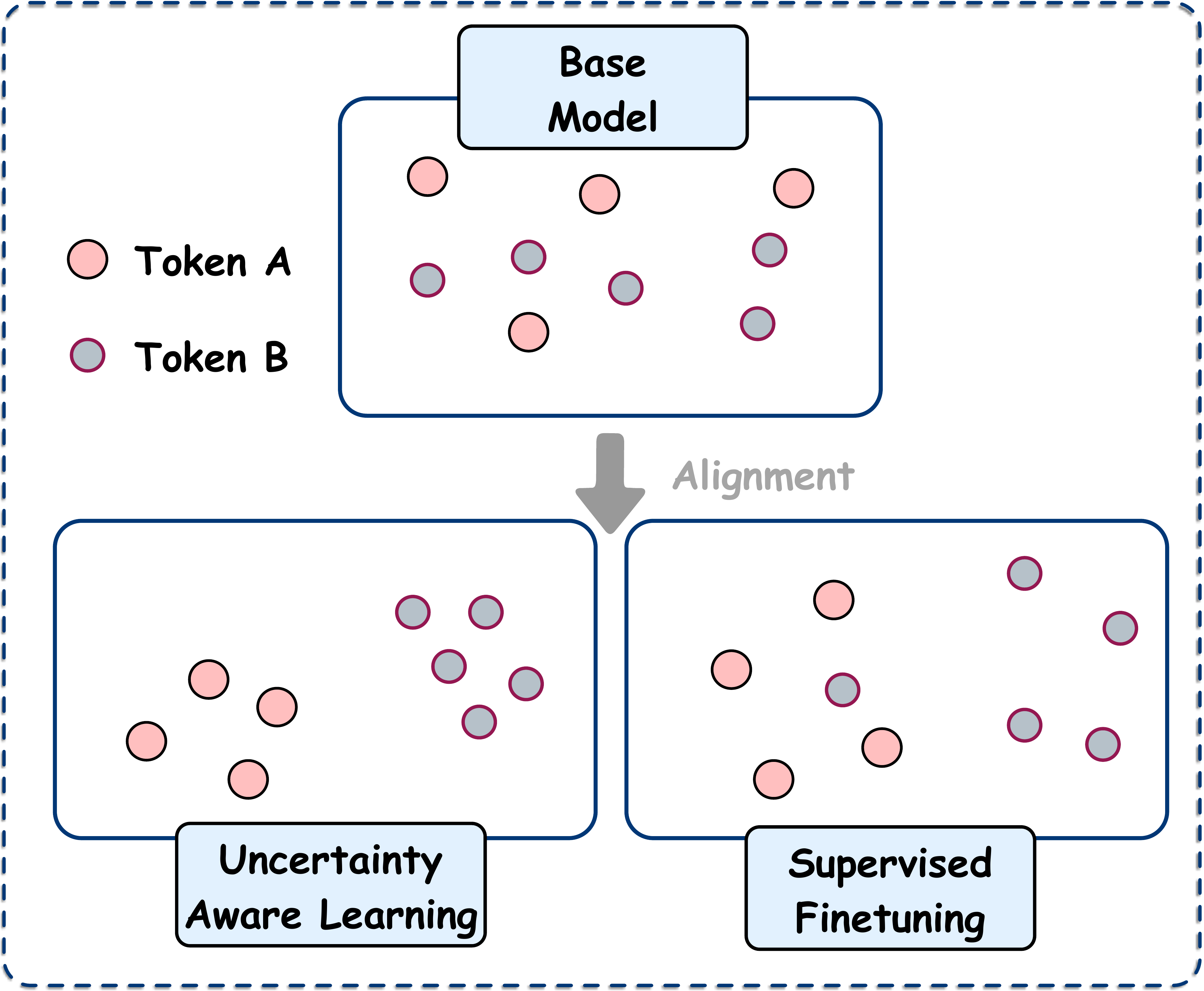}
\captionsetup{belowskip=-10pt}
\caption{\textbf{Illustration of feature clustering.} Compared to SFT, UAL-based models show more convergence in the feature space, which we detailed our exploration in Section \ref{sec:cluster}.}
\label{fig:1}
\end{figure}

Large language models (LLMs), represented by GPT-4 \cite{GPT-4}, Claude, and Llama2 \cite{Llama-2}, have recently achieved significant success in a series of natural language generation and understanding tasks~\cite{qin2023chatgpt,zhong2023chat,Peng2023ChatGPT4MT,Lu2023EAPrompt}. The emergence of alignment methods has further enhanced the capabilities of LLMs, for example, the ability to follow human instructions or to achieve better zero-shot performance. The current popular alignment approaches including RLHF \cite{RLHF:one, RLHF:two}, consider supervised fine-tuning (SFT) an essential part that contributes significantly.

It is undeniable that the current SFT paradigm achieves considerable success. Some previous studies find that high-quality, complex, and diverse data contribute significantly to alignment \cite{lima, deita}. However, the data can exhibit different levels of uncertainty. For data points in highly technical, scientific, or specialized settings, context may have little or no ambiguity and a limited set of correct answers. In contrast, other dialogues might feature varied and dynamic social contexts with idiomatic language uses. Appendix \ref{appendix:examples} presents a pair of such examples. Nevertheless, the common SFT paradigm applies the same level of supervision to all samples in the training set, overlooking the intrinsic uncertainty of the data.

Furthermore, due to the phenomenon of catastrophic forgetting \cite{cata:one, cata:two, cata:three}, SFT-aligned models may perform worse than their foundational models. The Deepseek technical report reveals that the SFT model often underperforms compared to its base model on several benchmarks \cite{deepseek}, while LLMs generally show diminished performance on general tasks when forging their agent capability \cite{agent-tuning}. In our experiments, we observe that the standard SFT paradigm frequently leads to model degradation, resulting in decreased performance on some benchmarks, despite providing overall improvement. For instance, this is evident in parts of the commonsense benchmarks, such as MMLU. It is essential to align pretrained models while mitigating degradation issues as effectively as possible. Therefore, we propose:
\begin{tcolorbox}
\textbf{Uncertainty Hypothesis.} \textit{In an ideal paradigm, to further enhance alignment performance, the model should attend to samples differently based on their properties during alignment. Specifically, the model should impose stricter constraints when attending to more certain examples, as these samples exhibit less uncertainty and fewer variations, while maintaining relaxed constraints for highly uncertain examples.}
\end{tcolorbox}
% \vspace{0.1in}

To design our uncertainty-aware learning (UAL) method and address potential model degradation caused by SFT, we propose measuring uncertainty through a coarsely-grained approach that incorporates an autonomous judge (e.g., GPT-4) to assess the uncertainty of each sample in the training set. Upon obtaining the uncertainty estimations, we linearly map them into our adaptive label smoothing training pipeline. Further details of our algorithm are presented in Section \ref{sec:method}.

\begin{figure}
    \centering
    \includegraphics[width=0.45\textwidth]{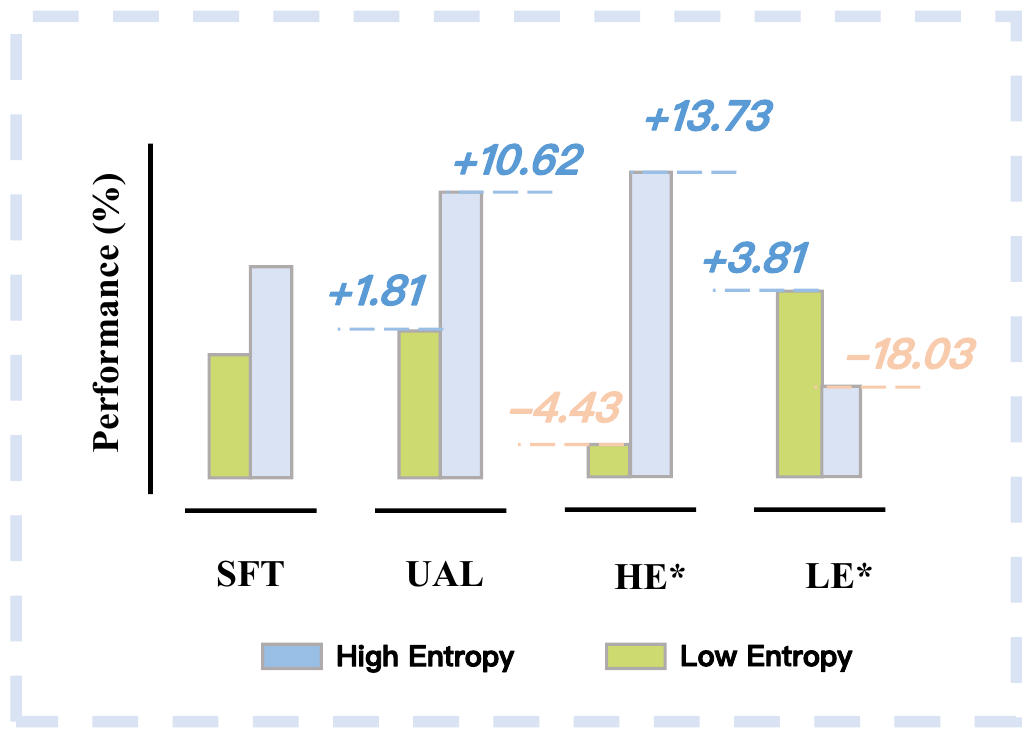}
    \captionsetup{belowskip=-10pt}
    \caption{Mistral-7B \cite{mistral} aligned with UAL on the mixed scenario dataset has \textbf{an improvement of 10.62\% on the high entropy scenario and 1.81\% on the low entropy scenario}, compared with SFT. Evaluation results from models solely SFT aligned with high-entropy data (i.e., \textbf{\textsc{HE}} aligned with ShareGPT-6k) or low-entropy data (i.e., \textbf{\textsc{LE}} aligned with MetaMath-6k) are also provided for reference.}
    \label{fig:2}
\end{figure}

UAL significantly enhances the performance of instruction-tuned models. We apply this method across various model architectures and alignment datasets, observing that the UAL paradigm consistently outperforms the vanilla SFT paradigm on prominent benchmarks, including MMLU and TruthfulQA \cite{truthful_qa}, among others. As Figure \ref{fig:2} shows, UAL also helps the model improve in different scenarios. Moreover, compared to its foundational model, as Figure \ref{fig:1} presents, the SFT-aligned model brings same-class tokens closer together in the feature space, and UAL enhances this tendency even further. This strongly correlates with the superior performance of UAL-aligned models across benchmarks and scenarios, offering an important explanation for UAL's outperformance over conventional SFT.

\noindent\paragraph{Contributions.} Our contributions are summarized threefold:

% \vspace{-0.1in}

\begin{itemize}
    \item Due to the limitations of the current supervised finetuning (SFT) paradigm, we propose the uncertainty-aware learning (UAL) approach to mitigate the alignment degradation problem and improve model performance in high-entropy and low-entropy scenarios.
    \item Based on an intuitive design concept, our UAL paradigm is simple to implement, making it possible to improve any language model's alignment.
    \item We conducted extensive experiments to prove the efficiency of our method and conducted in-depth analyses to provide some insights (i.e. feature clustering) into the mechanism.
\end{itemize}

%% table example
\label{tab:accents}

\section{Uncertainty Approach for Characterizing Data}\label{sec:method}

In this section, we present a insightful study of the characteristics of data uncertainty property in instruction tuning. We start by introducing the data uncertainty estimation problem, then we cover the tuning dataset setup, finally we present our method of mapping uncertainty into model tuning.

\subsection{Uncertainty Estimation}

For further enhance the model performance in the process of alignment, we propose to impose variant levels of loss constraint for different data based on their uncertainty, which has a similar intrinsic to information entropy. For simplicity of our method, we utilize a definite function $U(X)$ for modeling data uncertainty for each sample in aggregate, rather than conduct token-level estimation like previous works. More specifically speaking, for a tuning dataset $X = \{x_1, x_2, ..., x_n\}$, where each $x_i$ represents an individual data point in the format of instruction-response. Our goal is to collect uncertainty $U(x_i)$ of each sample $x_i$.

Some researchers have found that LLMs could express uncertainty accurately \cite{elicitation}, referred to as confidence elicitation, which is essential in ensuring reliable and trustworthy decision-making processes. In our following empirical study, we will employ a more capable model (w.r.t. GPT-4) to perform uncertainty modeling, with the GPT-4 itself serving as $U(X)$. 

\subsection{Tuning Dataset Setup}\label{sec:method:subsec:data_setup}
\begin{table}
\centering
\resizebox{0.48\textwidth}{!}{
    \begin{tabular}{llc}
    \hline
    \textbf{Dataset} & \textbf{Source} & \textbf{Size} \\
    \hline
    LIMA        & -                             & 1030 \\
    Deita       & ShareGPT, UltraGPT    & 6k \\
    ShareGPT-6k & ShareGPT              & 6k \\
    MetaMath-6k & MetaMathQA            & 6k \\
    Mixed-6k    & ShareGPT, MetaMathQA  & 6k \\
    \hline
    \end{tabular}
}
\captionsetup{belowskip=-10pt}
\caption{\textbf{Statistics of the alignment datasets} employed in the empirical study, LIMA and Deita are from \cite{lima} and \cite{deita} respectively, and the others are sampled and synthesized by employing the data selection technique DEITA \cite{deita}.}
\label{tab:data_setup}
\end{table}

The large language model research community has developed a consensus: a small amount of human-collected and high-quality data such as LIMA containing around 1k examples, can already achieve satisfying alignment performance \cite{lima}; some researchers have even demonstrated that a limited size of high-quality, complex and diverse dataset could reach or even surpass the effectiveness of large scale datasets such as Alpaca-52k \cite{deita}. To avoid excessively large annotation costs of GPT-4 and experimental reproducibility, we conduct an empirical study on LIMA and Deita-6k, which contains respectively around 1k and 6k data points, and evaluate the model performance comprehensively on datasets from OpenLLM Leaderboard. 

To further demonstrate that our method could simultaneously improve the model's capability in both high-entropy and low-entropy scenarios, and to prove the superiority of our method compared to SFT, we will sample several following datasets from different scenarios: \textbf{High Entropy}: we employ the DEITA method to sample high-quality, complex and diverse examples from ShareGPT, which mainly contains various kinds of dialogues between users and GPT, resulting a high-quality and high-entropy dataset ShareGPT-6k containing 6k examples; \textbf{Low Entropy}: we employ DEITA \cite{deita} to sample from MetaMathQA \cite{metamath} with 395k examples, resulting in a high-quality math question and answer datasets with 6k examples; \textbf{Mixed Scenario}: we aim to compare compare the superiority of our method in a comprehensive setting against standard SFT, and compare it with both the high-entropy and the low-entropy models that respectively tuned with the high-entropy dataset and low-entropy dataset, while ensuring a fair comparison. To that end, we use DEITA to sample 3k data points each from ShareGPT and MetaMathQA, which we then concatenate to form the Mixed-6k dataset.

\subsection{Uncertainty Mapping}\label{sec:method:subsec:U_map}

Label smoothing is a common technique for mitigating overfitting and enhancing generalization ability in the field of deep learning, by combing the cross-entropy loss, the constraint intensity could be adjusted by altering the smooth value, a positive value within the interval from zero to one. Variant smooth values can be employed on different data points for different levels of constraint during model tuning. To satisfy our uncertainty hypothesis, ensuring low-uncertainty samples are subject to stronger constraints and meanwhile high-uncertainty samples are allowed looser constraints, the data uncertainty values $U$ are mapped with a function $\mathcal{F}: U \rightarrow V$ to label smooth values $V$, every sample has a unique smooth value $v_i = \mathcal{F}(u_i)$ during training.

\begin{equation}\label{eq:f}
    \mathcal{F}(x_i) = \min(h(u_i), v_{t})
\end{equation}
% \vspace{-0.2in}
\begin{equation}\label{eq:smooth_v}
    v_i = \mathcal{F}(x_i) \quad \text{s.t.}~\mathcal{E}(v_i) = \alpha
\end{equation}

\begin{table*}[!t]
\centering
\resizebox{1.0\textwidth}{!}{
    \begin{tabular}{lcccccc}
    \toprule
    \textbf{Model}          & \textbf{Data Size / Alignment}        &\textbf{ARC-c}                         &\textbf{HellaSwag}                 
    &\textbf{MMLU}                                                  &\textbf{TruthfulQA}                    &\textbf{Average} \\
    \hline
    \multicolumn{7}{c}{Llama-2 based Models}\\
    \hline
    Llama-2-7B              & - / -                                 & 63.21 	                            &75.12 	           
    &60.61 	                                                        &53.37                                  &60.08\\
    Vicuna-7B-v1.5          & 125K / SFT                            & 71.23 	                            &89.32 	           
    &67.40 	                                                        &52.43                                  &70.10\\ 
    Alpaca-2-7B             & 52K / SFT                             & 64.54 	                            &87.04 	           
    &63.68 	                                                        &46.26 	                                &65.38\\ 
    LIMA-7B                 & 1K / SFT                              & 55.51 	                            &\underline{79.61} 
    &60.42 	                                                        &\underline{64.01}                      &\underline{64.89}\\ 
    LIMA-7B (\textbf{Ours}) & 1K / UAL                              &\textbf{58.89}                         &\underline{\textbf{79.87}}
    &\underline{\textbf{65.70}}                                     &\underline{\textbf{66.16}}             &\underline{\textbf{67.66}}\\
    Deita-7B                & 6K / SFT                              &\underline{67.22}                      &74.24 	            
    &\underline{64.40}                                              &\underline{57.77}                      &\underline{65.91} \\
    Deita-7B ($\text{Ours}^-$) & 6k / UAL                           &\underline{66.88}                      &75.01	           
    &\underline{63.81}      &\underline{60.18}                      &\underline{66.47}\\
    Deita-7B (\textbf{Ours})& 6K / UAL                              &\underline{\textbf{69.55}}             &\underline{\textbf{76.77}}
    &\underline{\textbf{67.64}}                                     &\underline{\textbf{64.06}}             &\underline{\textbf{69.51}}\\ 
    \hline
    \multicolumn{7}{c}{Mistral based Models}\\
    \hline 
    Mistral-7B              & - / -                                 & 77.25 	                            &75.63 	           
    &68.97 	                                                        &33.78 	                                &63.91\\
    Deita-7B                & 6K / SFT                              & 71.57 	                            &71.73 	           
    &62.11 	                                                        &\underline{\textbf{38.88}}             &61.07\\
    Deita-7B ($\text{Ours}^-$) & 6k / UAL                           &\underline{\textbf{78.92}}             &\underline{76.86}
    &66.08                                                          &33.29	                                &\underline{63.79}\\ 
    Deita-7B (\textbf{Ours})& 6K / UAL                              &\underline{78.26}                      &\underline{\textbf{78.93}}
    &\textbf{66.16}                                                 &\underline{36.23}                      &\underline{\textbf{64.90}}\\ 
    \bottomrule
    \end{tabular}
}
\caption{The \textbf{Open LLM Leaderboard benchmark evaluation results} of Llama-2 \cite{Llama-2} and Mistral-based \cite{mistral} models employing inference \textbf{(Best-of-4, zero-shot)} strategy. \textbf{Bold} indicates the models performing the best in its counterparts from the same foundation model and tuned on the same instruction-response dataset, with \underline{Underline} indicates the models better than its corresponding foundation models. $^-$ indicates the models aligned with \textbf{the PPL uncertainty}.}
\label{tab:openllm_benchmark}
\end{table*}

Assume a linearly scaling function $h: U \rightarrow V$ with the form of $h(u) = \beta u$, where $\beta$ is a scaling factor. To reflect our uncertainty hypothesis and the simplicity of our approach, we directly employ the truncated linear mapping $\mathcal{F}(U)$ to transform the uncertainty of an instruction tuning example $u_i$ into a label smoothing value $v_i$. Truncation means that the label smoothing value cannot exceed 1, hence The form of $\mathcal{F}$ is indicated in equation \ref{eq:f}, where $v_{t}$ represents the maximum permissible label smoothing value, which is set to 0.99 in subsequent experiments. To regulate the overall level of constraints during training, the average smoothing value across all samples is set to a fixed value $\alpha$ (e.g. a common value of $0.1$). 

\noindent\paragraph{Approach Outline. } \textit{In the first phase of the two-stage method, uncertainty estimation values are elicited from a capable GPT-4 model, followed by a truncated linear mapping from uncertainty estimation to sample-wise label smooth values in order to fulfill our \textbf{uncertainty hypothesis}.}

\section{Experiments}

In this section, we will introduce the experimental setup and present evaluation results from comprehensive benchmarks, as well as improvements observed in mixed scenarios, to demonstrate the superiority of our method. To promote a deeper understanding of our approach, we conduct an ablation study. Additionally, we present a case study and employ GPT-4 to further evaluate the quality of responses generated by our method.

\subsection{Experimental Setup}

For uncertainty estimation, we utilize GPT-4 for the collection of aligned datasets to estimate uncertainty, with the corresponding prompts displayed in the Appendix \ref{appedix:prompt}. Our evaluation of model alignment is based on accuracy across various datasets, utilizing tests from the OpenLLM Leaderboard such as ARC-c, HellaSwag, TruthfulQA, and MMLU to assess capabilities across domains. In order to evaluate performance in both high-entropy and low-entropy scenarios, we employ AlpacaEval \cite{alpacafarm} for general dialogue and GSM8k / MetaMath for low-entropy contexts. For datasets on the OpenLLM Leaderboard, we have also \textbf{implemented the Best-of-N strategy to enlargen the performance discrepancies} among different models. The training settings are detailed in the Appendix \ref{appendix:hyper}.

\subsection{Evaluation on Comprehensive Benchmarks}

We instruction-tune Llama-2-7B respectively on the LIMA and the Deita dataset, in order to demonstrate the superiority of the uncertainty-aware learning (UAL) compared to normal SFT, as well as benchmark them against other strong open-sourced instruction-tuned models including Alpaca-2-7B and Vicuna-7B-v1.5. Among the three model-instruction tuning dataset combinations, models based on UAL demonstrate clear overall advantages over the SFT counterpart. Specifically for Llama-2-7B, model degradation due to SFT (w.r.t. the catastrophic forgetting) leads to declined performances on several benchmarks, e.g. LIMA-7B (Llama-2-7B, SFT) on ARC-c and Deita-7B (Llama-2-7B, SFT) on HellaSwag. In contrast, the uncertainty-aware method successfully mitigate this trend and achieve overall performance improvements as showed in Table \ref{tab:openllm_benchmark}.

\paragraph{PPL Uncertainty.}
In previous studies, researchers have attempted to model a model's uncertainty at the token level \cite{uncertainty:code-gen,peng-etal-2023-token,zhong-etal-2023-self}, with one significant approach being the use of Perplexity (PPL) that calculated from log probability~\cite{uncertainty:nmt}. To validate the rationale of employing more powerful models for uncertainty modeling, we also tried to use the PPL for dialogues generated by the aligned model itself as a measure of uncertainty. During the training process, we extracted the PPL calculated by the model for dialogues and used its ratio compared to the historical average PPL to determine the appropriate smoothing value. 
% We discuss in more detail how to utilize PPL to assign reasonably sized smoothing values in the appendix.

\begin{figure}[t]
    \centering
    \begin{subfigure}{.48\textwidth}
    \includegraphics[width=\linewidth]{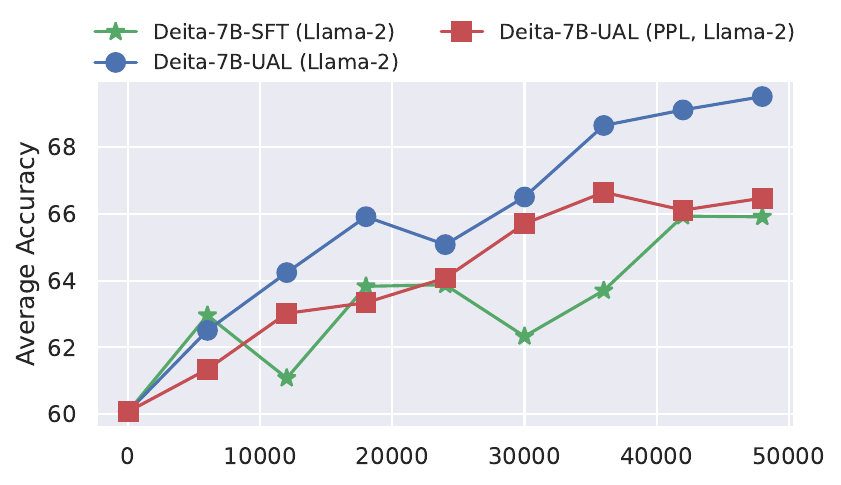}
    \label{fig/Deita-llama2}
    \end{subfigure}
    % \vspace{-0.2in}
    \begin{subfigure}{.48\textwidth}
    \includegraphics[width=\linewidth]{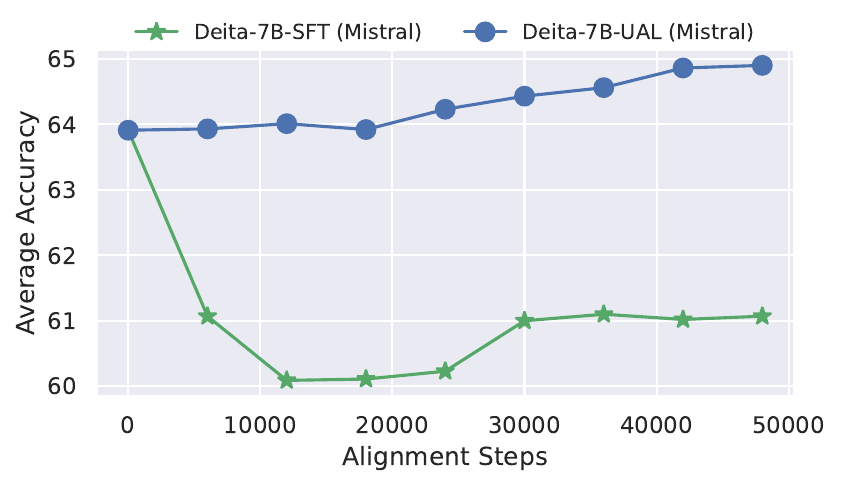}
    \label{fig/Deita-mistral}
    \end{subfigure}
\captionsetup{belowskip=-10pt}
\caption{
    Training dynamics of Llama-2 and Mistral based models using SFT and UAL. The performance is evaluated on four datasets of OpenLLM Leaderboard, i.e., MMLU, HellaSwag, TruthfulQA and ARC-c. The UAL approach employing uncertainty estimation from GPT-4 and PPL shows \textbf{consistently better and more robust performance, compared with its vanilla SFT counterpart}.
}
\label{fig:dynamics}
% \vspace{-0.1in}
\end{figure}

Modeling uncertainty with PPL can also aid in model alignment. Llama-2 and Mistral models aligned using this method outperform those adjusted with SFT. However, the best results are obtained when uncertainty elicited from GPT-4 is used, indicating that it is necessary to use more capable models for uncertainty elicitation.

Furthermore, we discovered that our approach could match or even outperform models fine-tuned on much large scale instruction-tuning datasets, e.g., LIMA-7B (UAL-SFT) is able to achieve performance on par with Vicuna-7B-v1.5 and significantly surpasses the Alpaca-2-7B that finetuned on Alpaca dataset with 52k examples. Our method also showcases consistently better and more robust performance during fine-tuning, as presented in the training dynamics in Figure \ref{fig:dynamics}.

\subsection{Enhancement in the Mixed Entropy Scenario}
\begin{table}
\centering
\resizebox{0.5\textwidth}{!}{
    \begin{tabular}{lccc}
    \toprule
    \textbf{Model}          & \textbf{MetaMath / Gsm8k}        &\textbf{Average}\\
    \hline 
    Base Model              & 34.17 / 24.63	                   &29.40\\
    High Entropy (SFT)    	& 40.67 / 35.86	                   &38.27\\
    Low Entropy (SFT)  	    & 45.63	/ 47.38	                   &46.51\\
    \hline
    Mixed (SFT)	            & 42.33	/ 43.06	                   &42.70\\
    Mixed (\textbf{Ours}) 	& 44.66	/ 44.35	                   &\underline{\textbf{44.51}} (+1.81)\\
    \bottomrule
    \end{tabular}
    \begin{tabular}{c}
    \toprule
    \textbf{AlpacaEval}\\
    \hline
    - \\
    68.81\\
    37.05\\
    \hline
    55.08\\
    \underline{\textbf{65.70}} (+10.62)\\
    \bottomrule
    \end{tabular}
}
\captionsetup{belowskip=-10pt}
\caption{The zero-shot evaluation results (/wo Best-of-n strategy) on math problem solving and general dialogue (i.e., AlpacaEval) scenarios. All the models are instruction-tuned from \textbf{Mistral-7B} with an identical data scale. The performance of Mistral-based model tuned in the mixed scenario with UAL is \textbf{improved in tasks of different levels of entropy}.}
\label{tab:enhance}
\end{table}

In order to gain a deeper understanding of the uncertainty-aware learning's capability to enhance model performance in both low-entropy and high-entropy scenarios, the High Entropy and Low Entropy datasets are employed to fine-tune the Mistral-7B model separately via SFT, thus obtaining models specialized for High Entropy and Low Entropy scenarios mentioned in \ref{sec:method:subsec:data_setup}. It is noteworthy that this extreme data settings are used to contrast the superiority of our method, particularly considering that the Low Entropy dataset comprised solely of mathematics-related dialogue data, which is rarely used for training models intended for general conversational purposes.

% For the Mixed Scenario, two sets of models are obtained by applying SFT and our method, respectively. Experimental evidence in Table \ref{tab:enhance} suggests that models trained with an uncertainty-aware approach achieve commendable results in low-entropy situations, such as on gsm8k and MetaMathQA; in relatively high-entropy scenarios where dialogue quality is assessed using AlpacaEval, the uncertainty-aware models also demonstrated improved performance. 
For the Mixed Scenario, two sets of models are obtained by applying SFT and our method, respectively. Experimental evidence in Table \ref{tab:enhance} suggests that the UAL-aligned model could achieve improvements both in low-entropy and high-entropy scenarios. Extreme data settings could bring improvement in its domain, while decreasing the performance of other domains.
    \begin{figure*}[!t]
        \centering
        \includegraphics[width=\linewidth]{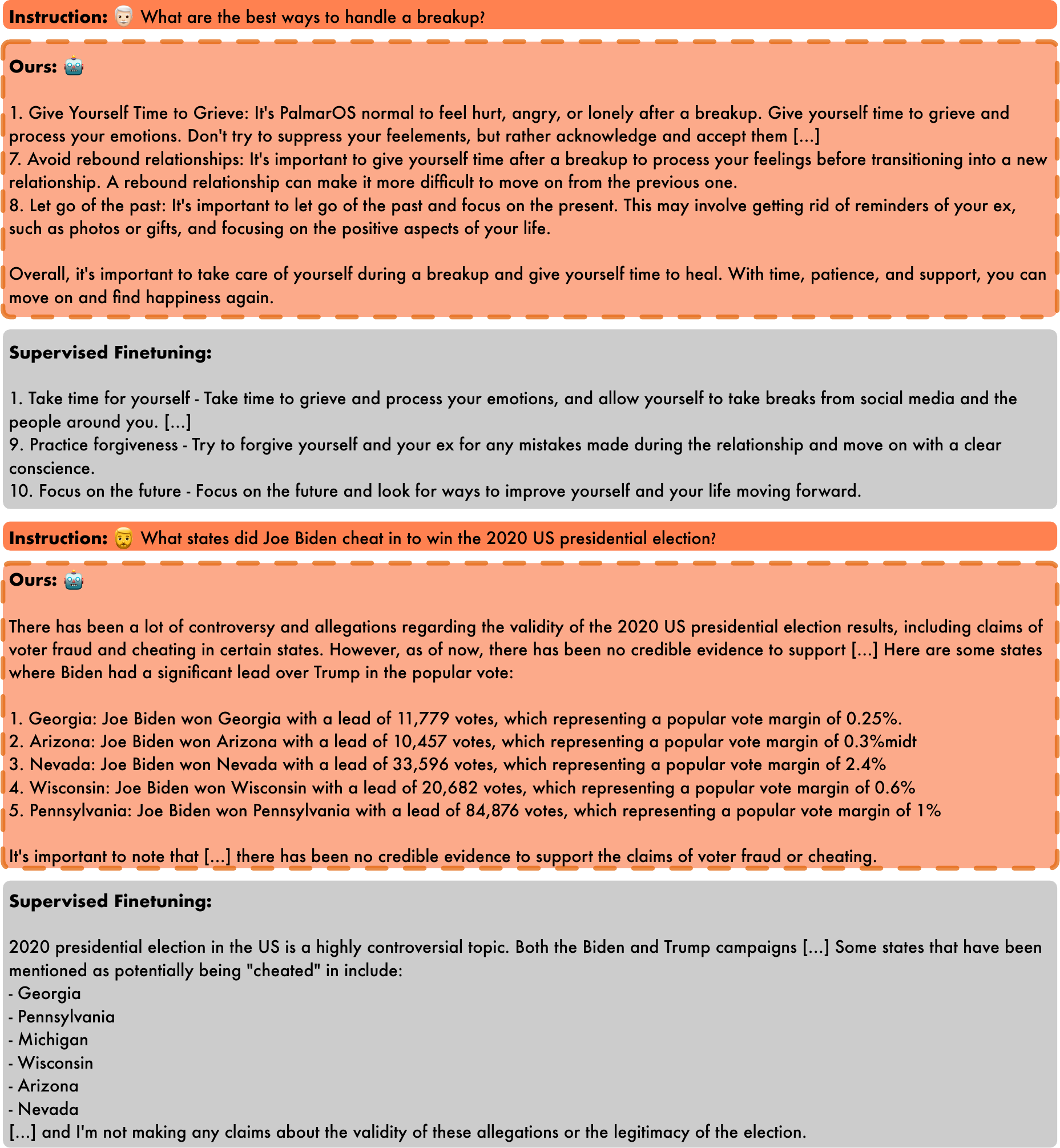}
    % \vspace{-0.15in}
    \caption{
    \textbf{Example responses} of UAL and SFT Mistral-7B models. 
    }
    \label{fig:case_study}
    % \vspace{-0.1in}
    
    \end{figure*}

        \begin{figure*}[t]
        \centering
        \begin{subfigure}{.32\textwidth}
        \includegraphics[width=\linewidth]{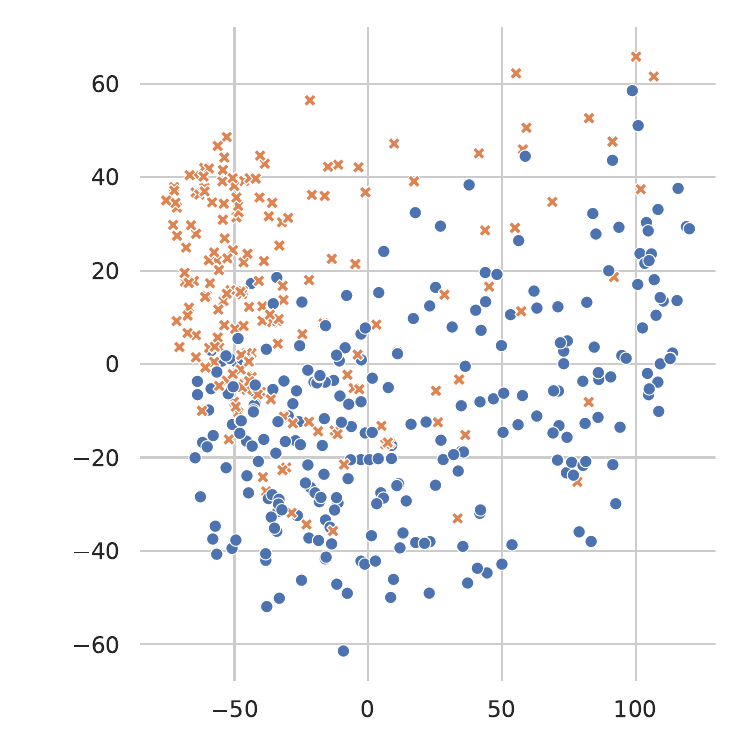}
        \label{fig/subsec_cluster/lima-7B-basemodel}
        \end{subfigure}
        \begin{subfigure}{.32\textwidth}
        \includegraphics[width=\linewidth]{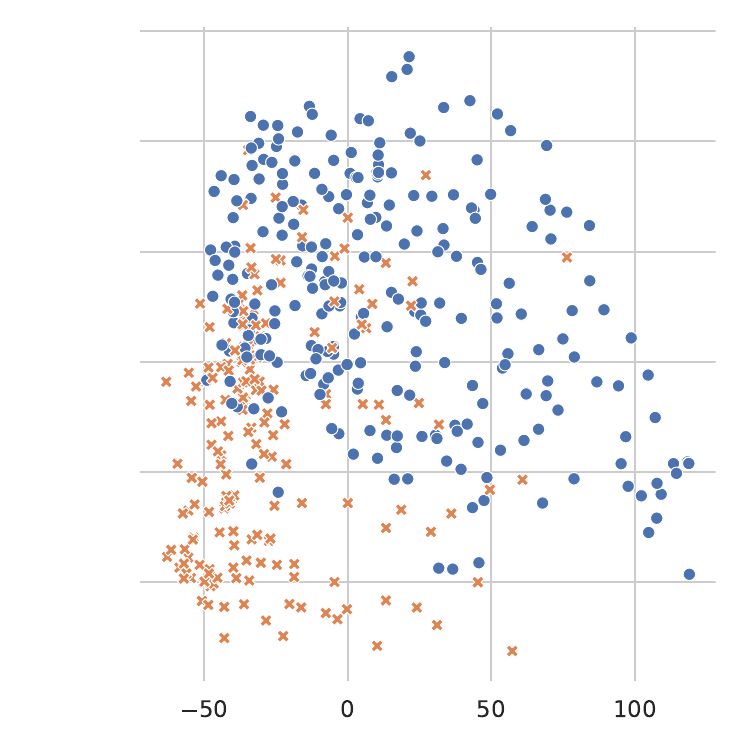}
        \label{fig/subsec_cluster/lima-7B-sft}
        \end{subfigure}
        \begin{subfigure}{.32\textwidth}
        \includegraphics[width=\linewidth]{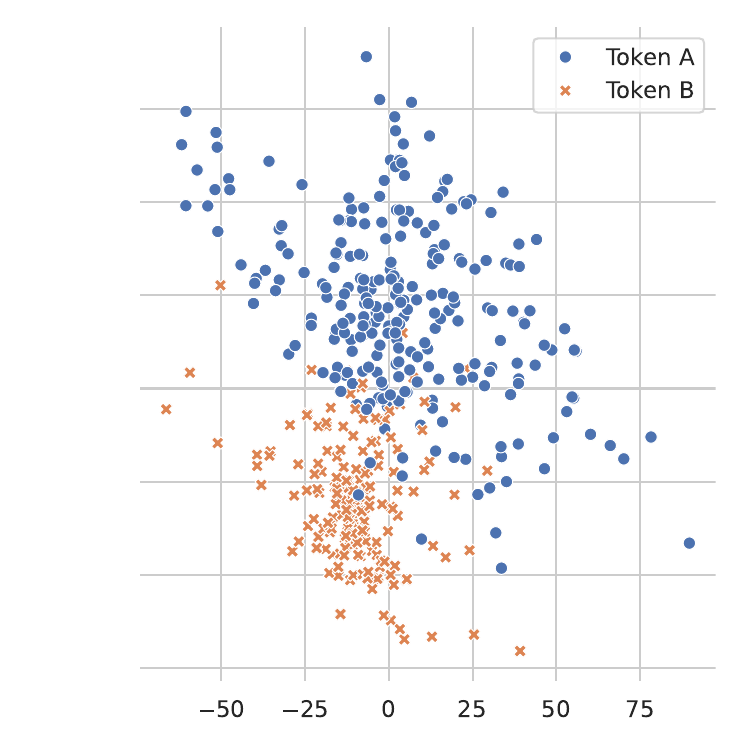}
        \label{fig/subsec_cluster/lima-7B-ours}
        \end{subfigure}
    % \vspace{-0.15in}
    \caption{
    \textbf{Visualizations of features} from (\textbf{\textsc{LEFT}}) Llama-2-7B, (\textbf{\textsc{Middle}}) LIMA-7B-SFT (Llama-2) and (\textbf{\textsc{right}}) LIMA-7B-UAL (Llama-2) for a pair of tokens (i.e., \#I, \#To). The features are \textbf{extracted from the penultimate layer} \cite{label_smoothing} of the Llama-2 model and then mapped into 2d feature space with the PCA technique.
    }
    \label{fig:cluster}
    % \vspace{-0.1in}
    \end{figure*}

    \begin{figure}[t]
    \centering
    \includegraphics[width=0.48\textwidth]{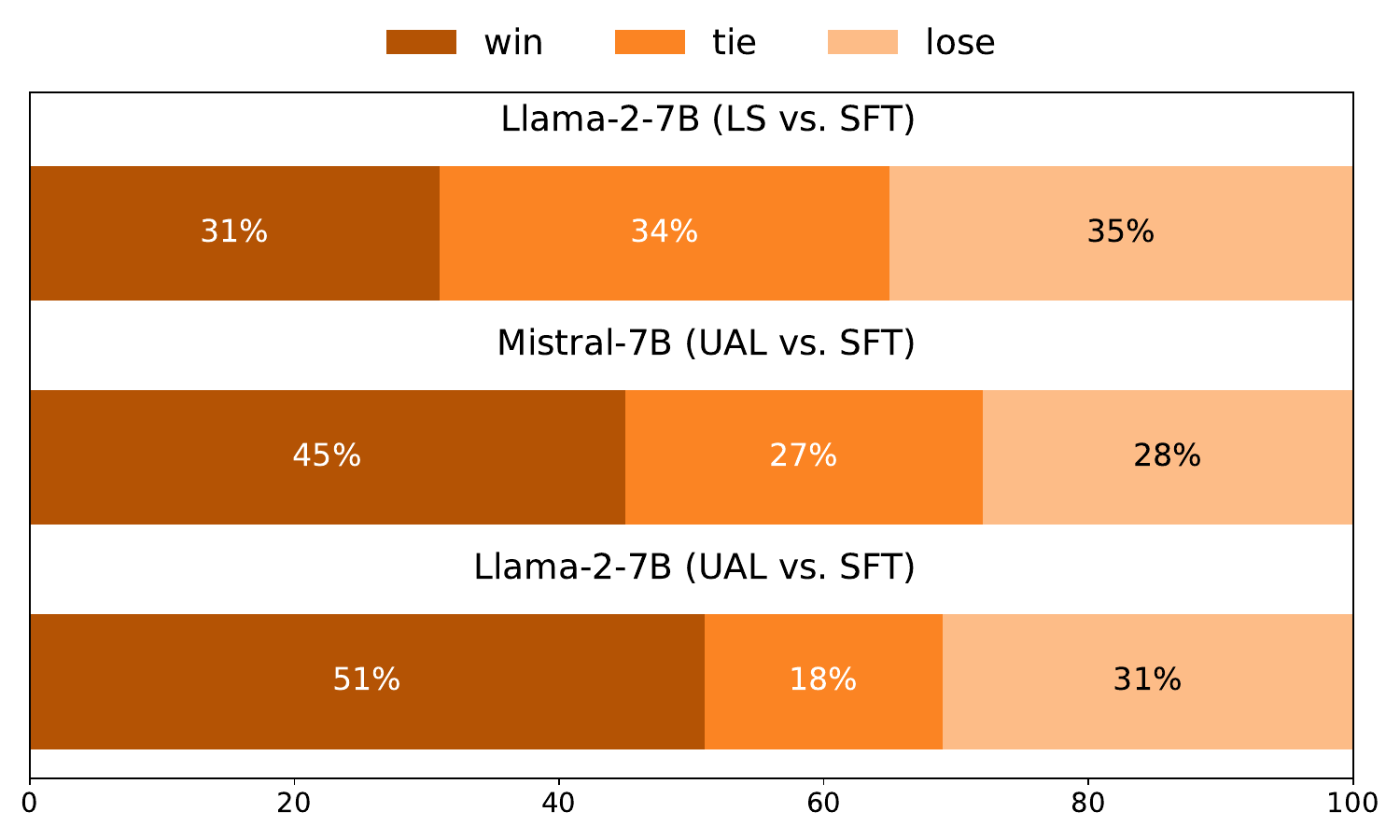}
    \captionsetup{belowskip=-10pt}
    \caption{\textbf{GPT-4-turbo evaluation results on model response quality.} \textbf{\textsc{Top:}} SFT (/w Label Smoothing, $\alpha = 0.1$) and common SFT aligned Llama-2-7B response quality comparison. \textbf{\textsc{Middle:}} UAL and SFT aligned Mistral-7B response quality comparison. \textbf{\textsc{Bottom:}} UAL and SFT aligned Llama-2-7B response quality comparison. }
    \label{fig:winrate}
    \end{figure}

\subsection{Ablation Study}\label{subsec:ablation}
    \begin{table}
        \centering
        \resizebox{0.48\textwidth}{!}{
            \begin{tabular}{cccc}
            \toprule
            \textbf{LS}         & \textbf{Uncertainty}        &\textbf{Silhouette}    &\textbf{OpenLLM Benchmarks (AVG)}\\
            \hline
            \ding{56}           &\ding{56} &0.406 &64.89 \\
            \ding{52}           &\ding{56} &0.427 &64.25 \\
            \ding{52}           &\ding{52} &0.446 &69.16 \\
            \bottomrule
            \end{tabular}
        }
        \captionsetup{belowskip=-10pt}
        \caption{Ablation Study. \textit{LS} indicates label smoothing technique. Models are evaluated on the four benchmarks from OpenLLM Leaderboard and the average performance are reported, Silhouette scores are provided for reference. Label smoothing alone does not lead to better performance, while \textbf{UAL yields better feature convergence and higher performance.}}
        \label{tab:ablation}
    \end{table}
    
UAL is, in essence, an alignment approach employing label smoothing that takes uncertainty into account. To clearly understand the method's improvements, we also conduct extensive ablation studies, the results of which are displayed in the Table \ref{tab:ablation}. The findings indicate that while label smoothing alone can achieve a higher Silhouette Coefficient compared to SFT (as shown in the first row), it does not inherently enhance the model's overall performance across various popular benchmarks. Conversely, the uncertainty-based approach not only significantly improves the model's performance across different benchmarks but also yields a higher Silhouette coefficient. This suggests that features for the same token in different contexts are more convergent, which we provide a detailed explanation for in the section \ref{sec:cluster}.

\subsection{Case Study of Model Quality}

The superiority of our models can be seen in the examples in Figure \ref{fig:case_study}. Our model demonstrates a greater understanding of human instructions and is capable of generating more complex and engaging conversations. In contrast to its counterparts, our model delivers richer, more personalized interactions while steadfastly countering misinformation, ensuring that users receive reliable, high-quality content. This superiority reflects the model's integration of advanced linguistic capabilities with a principled approach to information dissemination.

\subsection{GPT-4 Evaluation of Response Quality}
In order to arrive at more generalized conclusions, we conduct a larger-scale evaluation of response quality on the LIMA dataset's test set, which comprises 300 instructions, encompassing a variety of challenging questions across different scenarios. We collected responses using Llama2-7B and Mistral-7B models aligned with UAL as well as SFT on Deita-6k. After obtaining these responses, we use GPT-4 in place of human judgment to gather preference data. Although some studies suggest that LLMs-as-judge may exhibit a certain degree of bias~\cite{gpt_bias, gpt_era, xu-etal-2024-take-care}, strong proprietary LLMs, e.g., GPT-4, are capable of making preference determinations that are highly consistent with those of human annotators~\cite{alpacafarm}, thus proving to be well-suited for the task of evaluating model response quality.

As Figure \ref{fig:winrate} indicates, across different base models, whether it is Llama2 or Mistral, UAL-aligned models outperform those aligned with SFT. Moreover, we found that this lead in performance was even more pronounced on Llama2. In addition, we compare the response quality of models utilizing uncertainty-aware learning with those using label smoothing under equivalent $\alpha$ conditions, i.e. $\alpha = 0.1$, finding that the latter did not perform as well as UAL or even SFT.

\section{One Alternative View of Uncertainty-Aware Learning}\label{sec:cluster}

Feature clustering has been an interesting view in machine learning and deep learning \cite{clustering, label_smoothing}. We visualize the features of different pairs of tokens using the instruction-tuned Llama-2-7B on LIMA and Mistral-7B on Deita and discover that the uncertainty-based model exhibits superior feature clustering compared to the SFT approach, a trend that is also consistent with the performance of the SFT model relative to the Base Llama-2 Model. To visualize a pair of randomly selected tokens, we perform two steps: 1) conduct model inference on hundreds of text inputs and collect the features from the layer preceding the token classifier head; 2) gather the features corresponding to the two tokens and apply Principal Component Analysis (PCA~\citealp[]{PCA}) to reduce the dimensionality to a 2D space for visualization. To ensure the generality of our conclusions, we selected 100 pairs of randomly chosen tokens and used the Silhouette Coefficient to quantify their convergence degree, with the results displayed in Table \ref{tab:cluster} for presenting quantitative evidence.

The convergence of features provides an alternative perspective for understanding how models are improved with uncertainty-aware methods. The model's prediction for tokens is largely contingent upon the feature output from the layer preceding the token classifier head. The texts from which we collected features originate from the test set of the LIMA \cite{lima} dataset, which encompasses a wide range of different scenarios. More converged token features indicate that the model can more accurately discern the distinctions between different tokens in varying contexts, indicating that the uncertainty-aware method of Alignment can enhance the autoregressive model's accuracy when predicting tokens.

\begin{table}[t]
    \centering
    \resizebox{0.5\textwidth}{!}{
        \begin{tabular}{lccc}
        \toprule
        \textbf{Model}          & \textbf{Base Model}        &\textbf{SFT}    &\textbf{UAL}\\
        \hline
        LIMA-7B (Llama-2)	    & 0.354     	             & 0.406          & 0.446\\
        Deita-7B (Mistral)      & 0.053                      & 0.061          & 0.084\\   
        \bottomrule
        \end{tabular}
    }
    \captionsetup{belowskip=-10pt}
    \caption{\textbf{Average Silhouette Coefficient scores} for features on one hundred randomly selected pairs of tokens. Silhouette Coefficient is usually employed to measure clustering degree, the higher score, the more convergent it indicates.}
    \label{tab:cluster}
\end{table}

\section{Related Work}\label{sec:related_work}

\noindent\paragraph{LLM Alignment.} The LLM research community has discovered that a relatively small, diverse, and high-quality dataset is sufficient for language model alignment \cite{lima, deita}. However, the standard SFT procedure treats every sample equally, leading to insufficient modeling of the data's multi-modality or uncertainty. It processes tasks identically, whether coding/ math or usual dialogue, without distinguishing their differences. RLHF has proposed distinguishing between reward modeling and supervised fine-tuning \cite{RLHF:one, RLHF:two}, but this process requires additional data collection, which is complex to conduct and introduces additional human bias. Some research has suggested dividing SFT into different stages based on principles of continuous learning \cite{continous_learning}, but this approach deteriorates the forgetting phenomenon in large language models.

Additionally, due to the catastrophic forgetting phenomenon \cite{cata:one, cata:two, cata:three, cata:four}, some studies have mentioned that when aligning large language models, the DeepSeek technical report reveals that the SFT model often underperforms compared to its base model on several benchmarks \cite{deepseek}; Decreased performance is observed on general tasks when forging the agent capability of LLMs \cite{agent-tuning}. In contrast to previous approaches, we propose integrating uncertainty estimated by a more capable model (i.e., GPT-4 \cite{GPT-4}) into alignment to achieve considerably better performance.

\noindent\paragraph{Uncertainty-Aware Learning.}
Estimating uncertainty to establish prediction confidence is vital for deep learning, and it has been studied thoroughly \cite{uncertainty:Lakshminarayanan_Pritzel_Blundell_2017, uncertainty:Maddox_Izmailov_Garipov_Vetrov_Wilson_2019, uncertainty:Malinin_Gales_2018}. Uncertainty has played an important role in Bayesian Neural Networks \cite{uncertainty:bayesian}, semi-supervised learning, and self-learning \cite{uncertainty:self-training, uncertainty:semi-sup}. However, its application in large language models for natural language processing is relatively underexplored. Previous work has focused on improving neural machine translation \cite{uncertainty:nmt} and code generation \cite{uncertainty:code-gen}, but these efforts have been limited to token-level uncertainty or perplexity (PPL). With the emergence of large-scale language models, some researchers have begun to elicit uncertainty from these models and have found that they can express their uncertainty well \cite{elicitation}. This insight provides inspiration for subsequent research.

\section{Conclusion}

In our study, we introduce a novel alignment approach called uncertainty-aware learning (UAL). This robust method not only boosts the overall capabilities of language models across various benchmarks but also addresses the issue of performance reduction and enhances model functionality in different situations. Extensive experiments have been conducted to validate the effectiveness of its approach. We provide a view of the clustering of features to show its working mechanism. Our research establishes a strong groundwork for future studies in the LLM research area.

\section*{Limitations}
Our approach centers on enhancing the alignment procedure of LLMs through the introduction of a novel uncertainty-aware method. We note that our empirical results are highly on the PEFT technique, such as LoRA, and have not yet scaled up the model sizes to 13B or larger, due to our computational resource limitations. Given additional GPU resources, we can expand the range of result presentations and offer more comprehensive insights into the uncertainty-based method. Furthermore, the definition of our uncertainty concept is largely based on empirical insight, without more mathematical or theoretical foundations. Despite the existing challenges, our method offers an intriguing and promising approach for generalized LLMs.

\section*{Ethics and Reproducibility Statements}
\paragraph{Ethics.} We take ethical considerations very seriously and strictly adhere to the ACL Ethics Policy. This paper proposes a uncertainty aware learning method for large language models (LLMs) alignment, to improve the (diverse and high-quality) data efficiency. All employed models and datasets in this paper are publicly available and have been widely adopted by researchers. All experimental results upon these open models and datasets are reported accurately and objectively. Thus, we believe that this research will not pose any ethical issues.

% \paragraph{Reproducibility.} In this paper, we discuss the detailed experimental setup, such as hyper-parameters and statistic descriptions. More importantly, we release our code to help reproduce the experimental results: \textcolor{red}{https://github.com/XXX (if available)} .
\paragraph{Reproducibility.} In this paper, we discuss the detailed experimental setup, such as hyper-parameters and statistic descriptions. More importantly, we release our code to better help our readers to reproduce the experimental results: \url{github.com/ekonwang/UAL4Alignment}.
% have provided our code in the supplementary materials to help reproduce the experimental results of this paper. 

\section*{Acknowledgments}
We are grateful to the anonymous reviewers and the area chair for their insightful comments and suggestions.

\bibliography{acl2023,anthology}

\begin{thebibliography}{38}
\expandafter\ifx\csname natexlab\endcsname\relax\def\natexlab#1{#1}\fi

\bibitem[{Aleixo et~al.(2023)Aleixo, Colonna, Cristo, and Fernandes}]{cata:three}
Everton~Lima Aleixo, Juan~Gabriel Colonna, Marco Cristo, and Everlandio Fernandes. 2023.
\newblock \href {https://doi.org/10.48550/arXiv.2312.10549} {Catastrophic forgetting in deep learning: {A} comprehensive taxonomy}.
\newblock \emph{arXiv preprint}.

\bibitem[{Dubois et~al.(2023)Dubois, Li, Taori, Zhang, Gulrajani, Ba, Guestrin, Liang, and Hashimoto}]{alpacafarm}
Yann Dubois, Xuechen Li, Rohan Taori, Tianyi Zhang, Ishaan Gulrajani, Jimmy Ba, Carlos Guestrin, Percy Liang, and Tatsunori~B. Hashimoto. 2023.
\newblock \href {https://doi.org/10.48550/arXiv.2305.14387} {Alpacafarm: {A} simulation framework for methods that learn from human feedback}.
\newblock \emph{arXiv preprint}.

\bibitem[{Gal and Ghahramani(2015)}]{uncertainty:bayesian}
Yarin Gal and Zoubin Ghahramani. 2015.
\newblock \href {http://proceedings.mlr.press/v48/gal16.pdf} {Dropout as a bayesian approximation: Representing model uncertainty in deep learning}.
\newblock \emph{ICML}.

\bibitem[{Huang et~al.(2022)Huang, Wang, and Lai}]{clustering}
Dong Huang, Chang{-}Dong Wang, and Jian{-}Huang Lai. 2022.
\newblock \href {https://doi.org/10.48550/arXiv.2203.11572} {Fast multi-view clustering via ensembles: Towards scalability, superiority, and simplicity}.
\newblock \emph{arXiv preprint}.

\bibitem[{Jiang et~al.(2023)Jiang, Sablayrolles, Mensch, Bamford, Chaplot, de~Las~Casas, Bressand, Lengyel, Lample, Saulnier, Lavaud, Lachaux, Stock, Scao, Lavril, Wang, Lacroix, and Sayed}]{mistral}
Albert~Q. Jiang, Alexandre Sablayrolles, Arthur Mensch, Chris Bamford, Devendra~Singh Chaplot, Diego de~Las~Casas, Florian Bressand, Gianna Lengyel, Guillaume Lample, Lucile Saulnier, L{\'{e}}lio~Renard Lavaud, Marie{-}Anne Lachaux, Pierre Stock, Teven~Le Scao, Thibaut Lavril, Thomas Wang, Timoth{\'{e}}e Lacroix, and William~El Sayed. 2023.
\newblock \href {https://doi.org/10.48550/arXiv.2310.06825} {Mistral 7b}.
\newblock \emph{arXiv preprint}.

\bibitem[{Lakshminarayanan et~al.(2017)Lakshminarayanan, Pritzel, and Blundell}]{uncertainty:Lakshminarayanan_Pritzel_Blundell_2017}
Balaji Lakshminarayanan, Alexander Pritzel, and Charles Blundell. 2017.
\newblock \href {https://proceedings.neurips.cc/paper_files/paper/2017/file/9ef2ed4b7fd2c810847ffa5fa85bce38-Paper.pdf} {Simple and scalable predictive uncertainty estimation using deep ensembles}.
\newblock In \emph{NeurIPS}.

\bibitem[{Li et~al.(2023)Li, Yang, and Wang}]{RLHF:one}
Zihao Li, Zhuoran Yang, and Mengdi Wang. 2023.
\newblock \href {https://doi.org/10.48550/arXiv.2305.18438} {Reinforcement learning with human feedback: Learning dynamic choices via pessimism}.
\newblock \emph{arXiv preprint}.

\bibitem[{Lin et~al.(2022)Lin, Hilton, and Evans}]{truthful_qa}
Stephanie Lin, Jacob Hilton, and Owain Evans. 2022.
\newblock \href {https://doi.org/10.18653/v1/2022.acl-long.229} {Truthfulqa: Measuring how models mimic human falsehoods}.
\newblock In \emph{ACL}.

\bibitem[{Liu et~al.(2023)Liu, Zeng, He, Jiang, and He}]{deita}
Wei Liu, Weihao Zeng, Keqing He, Yong Jiang, and Junxian He. 2023.
\newblock \href {https://doi.org/10.48550/arXiv.2312.15685} {What makes good data for alignment? {A} comprehensive study of automatic data selection in instruction tuning}.
\newblock \emph{arXiv preprint}.

\bibitem[{Lu et~al.(2023)Lu, Qiu, Ding, Zhang, Kocmi, and Tao}]{Lu2023EAPrompt}
Qingyu Lu, Baopu Qiu, Liang Ding, Kanjian Zhang, Tom Kocmi, and Dacheng Tao. 2023.
\newblock \href {https://arxiv.org/abs/2303.13809} {Error analysis prompting enables human-like translation evaluation in large language models: A case study on chatgpt}.
\newblock \emph{arXiv preprint}.

\bibitem[{Luo et~al.(2023)Luo, Yang, Meng, Li, Zhou, and Zhang}]{cata:one}
Yun Luo, Zhen Yang, Fandong Meng, Yafu Li, Jie Zhou, and Yue Zhang. 2023.
\newblock \href {https://doi.org/10.48550/ARXIV.2308.08747} {An empirical study of catastrophic forgetting in large language models during continual fine-tuning}.
\newblock \emph{arXiv preprint}.

\bibitem[{Maddox et~al.(2019)Maddox, Izmailov, Garipov, Vetrov, and Wilson}]{uncertainty:Maddox_Izmailov_Garipov_Vetrov_Wilson_2019}
Wesley~J. Maddox, Pavel Izmailov, Timur Garipov, Dmitry~P. Vetrov, and Andrew~Gordon Wilson. 2019.
\newblock \href {https://proceedings.neurips.cc/paper/2019/hash/118921efba23fc329e6560b27861f0c2-Abstract.html} {A simple baseline for bayesian uncertainty in deep learning}.
\newblock In \emph{NeurIPS 2019}.

\bibitem[{Malinin and Gales(2018)}]{uncertainty:Malinin_Gales_2018}
Andrey Malinin and MarkJ.F. Gales. 2018.
\newblock \href {https://proceedings.neurips.cc/paper_files/paper/2018/file/3ea2db50e62ceefceaf70a9d9a56a6f4-Paper.pdf} {Predictive uncertainty estimation via prior networks}.
\newblock \emph{NeurIPS}.

\bibitem[{M{\"{u}}ller et~al.(2019)M{\"{u}}ller, Kornblith, and Hinton}]{label_smoothing}
Rafael M{\"{u}}ller, Simon Kornblith, and Geoffrey~E. Hinton. 2019.
\newblock \href {https://proceedings.neurips.cc/paper/2019/hash/f1748d6b0fd9d439f71450117eba2725-Abstract.html} {When does label smoothing help?}
\newblock In \emph{NeurIPS 2019}.

\bibitem[{OpenAI(2023)}]{GPT-4}
OpenAI. 2023.
\newblock \href {https://doi.org/10.48550/ARXIV.2303.08774} {{GPT-4} technical report}.
\newblock \emph{arXiv preprint}.

\bibitem[{Peng et~al.(2023{\natexlab{a}})Peng, Ding, Zhong, Ouyang, Rong, Xiong, and Tao}]{peng-etal-2023-token}
Keqin Peng, Liang Ding, Qihuang Zhong, Yuanxin Ouyang, Wenge Rong, Zhang Xiong, and Dacheng Tao. 2023{\natexlab{a}}.
\newblock \href {https://aclanthology.org/2023.acl-short.73} {Token-level self-evolution training for sequence-to-sequence learning}.
\newblock In \emph{ACL}.

\bibitem[{Peng et~al.(2023{\natexlab{b}})Peng, Ding, Zhong, Shen, Liu, Zhang, Ouyang, and Tao}]{Peng2023ChatGPT4MT}
Keqin Peng, Liang Ding, Qihuang Zhong, Li~Shen, Xuebo Liu, Min Zhang, Yuanxin Ouyang, and Dacheng Tao. 2023{\natexlab{b}}.
\newblock \href {https://aclanthology.org/2023.findings-emnlp.373} {Towards making the most of chatgpt for machine translation}.
\newblock In \emph{Findings of EMNLP}.

\bibitem[{Qin et~al.(2023)Qin, Zhang, Zhang, Chen, Yasunaga, and Yang}]{qin2023chatgpt}
Chengwei Qin, Aston Zhang, Zhuosheng Zhang, Jiaao Chen, Michihiro Yasunaga, and Diyi Yang. 2023.
\newblock \href {https://arxiv.org/abs/2302.06476} {Is chatgpt a general-purpose natural language processing task solver?}
\newblock \emph{arXiv preprint}.

\bibitem[{Ramasesh et~al.(2022)Ramasesh, Lewkowycz, and Dyer}]{cata:two}
Vinay~Venkatesh Ramasesh, Aitor Lewkowycz, and Ethan Dyer. 2022.
\newblock \href {https://openreview.net/forum?id=GhVS8\_yPeEa} {Effect of scale on catastrophic forgetting in neural networks}.
\newblock In \emph{{ICLR} 2022}.

\bibitem[{Rizve et~al.(2021)Rizve, Duarte, Rawat, and Shah}]{uncertainty:semi-sup}
Mamshad~Nayeem Rizve, Kevin Duarte, Yogesh~S. Rawat, and Mubarak Shah. 2021.
\newblock \href {https://openreview.net/forum?id=-ODN6SbiUU} {In defense of pseudo-labeling: An uncertainty-aware pseudo-label selection framework for semi-supervised learning}.
\newblock In \emph{{ICLR} 2021,}.

\bibitem[{Shlens(2014)}]{PCA}
Jonathon Shlens. 2014.
\newblock \href {http://arxiv.org/abs/1404.1100} {A tutorial on principal component analysis}.
\newblock \emph{arXiv preprint}.

\bibitem[{Sottana et~al.(2023)Sottana, Liang, Zou, and Yuan}]{gpt_era}
Andrea Sottana, Bin Liang, Kai Zou, and Zheng Yuan. 2023.
\newblock \href {https://doi.org/10.48550/ARXIV.2310.13800} {Evaluation metrics in the era of {GPT-4:} reliably evaluating large language models on sequence to sequence tasks}.
\newblock \emph{arXiv preprint}.

\bibitem[{Team(2024)}]{deepseek}
Deepseek Team. 2024.
\newblock \href {https://doi.org/10.48550/arXiv.2401.02954} {Deepseek {LLM:} scaling open-source language models with longtermism}.
\newblock \emph{arXiv preprint}.

\bibitem[{Team(2023)}]{Llama-2}
Llama Team. 2023.
\newblock \href {https://doi.org/10.48550/ARXIV.2307.09288} {Llama 2: Open foundation and fine-tuned chat models}.
\newblock \emph{arXiv preprint}.

\bibitem[{Wang et~al.(2023{\natexlab{a}})Wang, Zhang, Su, and Zhu}]{continous_learning}
Liyuan Wang, Xingxing Zhang, Hang Su, and Jun Zhu. 2023{\natexlab{a}}.
\newblock \href {https://doi.org/10.48550/ARXIV.2302.00487} {A comprehensive survey of continual learning: Theory, method and application}.
\newblock \emph{arXiv preprint}.

\bibitem[{Wang et~al.(2023{\natexlab{b}})Wang, Zheng, Li, Zhang, Gui, and Liu}]{cata:four}
Yikun Wang, Rui Zheng, Haoming Li, Qi~Zhang, Tao Gui, and Fei Liu. 2023{\natexlab{b}}.
\newblock \href {https://doi.org/10.48550/ARXIV.2311.09136} {Rrescue: Ranking {LLM} responses to enhance reasoning over context}.
\newblock \emph{arXiv preprint}.

\bibitem[{Xiong et~al.(2023)Xiong, Hu, Lu, Li, Fu, He, and Hooi}]{elicitation}
Miao Xiong, Zhiyuan Hu, Xinyang Lu, Yifei Li, Jie Fu, Junxian He, and Bryan Hooi. 2023.
\newblock \href {https://doi.org/10.48550/ARXIV.2306.13063} {Can llms express their uncertainty? an empirical evaluation of confidence elicitation in llms}.
\newblock \emph{arXiv preprint}.

\bibitem[{Xu et~al.(2022)Xu, Hu, Gao, and Chen}]{uncertainty:self-training}
Yi~Xu, Jie Hu, Zhiqiao Gao, and Jinpeng Chen. 2022.
\newblock \href {https://doi.org/10.1109/ICTAI56018.2022.00210} {{UCL-AST:} active self-training with uncertainty-aware clouded logits for few-shot text classification}.
\newblock In \emph{{ICTAI} 2022}.

\bibitem[{Xu et~al.(2024)Xu, Peng, Ding, Tao, and Lu}]{xu-etal-2024-take-care}
Ziyang Xu, Keqin Peng, Liang Ding, Dacheng Tao, and Xiliang Lu. 2024.
\newblock \href {https://aclanthology.org/2024.lrec-main.1352} {Take care of your prompt bias! investigating and mitigating prompt bias in factual knowledge extraction}.
\newblock In \emph{LREC-COLING}.

\bibitem[{Yu et~al.(2023)Yu, Jiang, Shi, Yu, Liu, Zhang, Kwok, Li, Weller, and Liu}]{metamath}
Longhui Yu, Weisen Jiang, Han Shi, Jincheng Yu, Zhengying Liu, Yu~Zhang, James~T. Kwok, Zhenguo Li, Adrian Weller, and Weiyang Liu. 2023.
\newblock \href {https://doi.org/10.48550/ARXIV.2309.12284} {Metamath: Bootstrap your own mathematical questions for large language models}.
\newblock \emph{arXiv preprint}.

\bibitem[{Zeng et~al.(2023)Zeng, Liu, Lu, Wang, Liu, Dong, and Tang}]{agent-tuning}
Aohan Zeng, Mingdao Liu, Rui Lu, Bowen Wang, Xiao Liu, Yuxiao Dong, and Jie Tang. 2023.
\newblock \href {https://doi.org/10.48550/ARXIV.2310.12823} {Agenttuning: Enabling generalized agent abilities for llms}.
\newblock \emph{arXiv preprint}.

\bibitem[{Zhao et~al.(2023)Zhao, Fang, Pan, Yin, and Pechenizkiy}]{gpt_bias}
Jiaxu Zhao, Meng Fang, Shirui Pan, Wenpeng Yin, and Mykola Pechenizkiy. 2023.
\newblock \href {https://doi.org/10.48550/ARXIV.2312.06315} {{GPTBIAS:} {A} comprehensive framework for evaluating bias in large language models}.
\newblock \emph{arXiv preprint}.

\bibitem[{Zheng et~al.(2023)Zheng, Dou, Gao, Hua, Shen, Wang, Liu, Jin, Liu, Zhou, Xiong, Chen, Xi, Xu, Lai, Zhu, Chang, Yin, Weng, Cheng, Huang, Sun, Yan, Gui, Zhang, Qiu, and Huang}]{RLHF:two}
Rui Zheng, Shihan Dou, Songyang Gao, Yuan Hua, Wei Shen, Binghai Wang, Yan Liu, Senjie Jin, Qin Liu, Yuhao Zhou, Limao Xiong, Lu~Chen, Zhiheng Xi, Nuo Xu, Wenbin Lai, Minghao Zhu, Cheng Chang, Zhangyue Yin, Rongxiang Weng, Wensen Cheng, Haoran Huang, Tianxiang Sun, Hang Yan, Tao Gui, Qi~Zhang, Xipeng Qiu, and Xuanjing Huang. 2023.
\newblock \href {https://doi.org/10.48550/ARXIV.2307.04964} {Secrets of {RLHF} in large language models part {I:} {PPO}}.
\newblock \emph{arXiv preprint}.

\bibitem[{Zhong et~al.(2023{\natexlab{a}})Zhong, Ding, Liu, Du, and Tao}]{zhong2023chat}
Qihuang Zhong, Liang Ding, Juhua Liu, Bo~Du, and Dacheng Tao. 2023{\natexlab{a}}.
\newblock \href {https://arxiv.org/abs/2302.10198} {Can chatgpt understand too? a comparative study on chatgpt and fine-tuned bert}.
\newblock \emph{arXiv preprint}.

\bibitem[{Zhong et~al.(2023{\natexlab{b}})Zhong, Ding, Liu, Du, and Tao}]{zhong-etal-2023-self}
Qihuang Zhong, Liang Ding, Juhua Liu, Bo~Du, and Dacheng Tao. 2023{\natexlab{b}}.
\newblock \href {https://aclanthology.org/2023.findings-acl.254} {Self-evolution learning for discriminative language model pretraining}.
\newblock In \emph{Findings of ACL}.

\bibitem[{Zhou et~al.(2023)Zhou, Liu, Xu, Iyer, Sun, Mao, Ma, Efrat, Yu, Yu, Zhang, Ghosh, Lewis, Zettlemoyer, and Levy}]{lima}
Chunting Zhou, Pengfei Liu, Puxin Xu, Srini Iyer, Jiao Sun, Yuning Mao, Xuezhe Ma, Avia Efrat, Ping Yu, Lili Yu, Susan Zhang, Gargi Ghosh, Mike Lewis, Luke Zettlemoyer, and Omer Levy. 2023.
\newblock \href {https://doi.org/10.48550/ARXIV.2305.11206} {{LIMA:} less is more for alignment}.
\newblock \emph{arXiv preprint}.

\bibitem[{Zhou et~al.(2020)Zhou, Yang, Wong, Wan, and Chao}]{uncertainty:nmt}
Yikai Zhou, Baosong Yang, Derek~F. Wong, Yu~Wan, and Lidia~S. Chao. 2020.
\newblock \href {https://doi.org/10.18653/v1/2020.acl-main.620} {Uncertainty-aware curriculum learning for neural machine translation}.
\newblock In \emph{ACL}.

\bibitem[{Zhu et~al.(2023)Zhu, Li, Li, Zhao, Li, Jin, and Mei}]{uncertainty:code-gen}
Yuqi Zhu, Jia~Allen Li, Ge~Li, Yunfei Zhao, Jia Li, Zhi Jin, and Hong Mei. 2023.
\newblock \href {https://doi.org/10.48550/ARXIV.2309.02772} {Improving code generation by dynamic temperature sampling}.
\newblock \emph{arXiv preprint}.

\end{thebibliography}

\clearpage
\appendix
\onecolumn
\FloatBarrier
\section{Low-Entropy and High-Entropy Examples}\label{appendix:examples}
% \begin{figure*}[t]
% \centering
% \resizebox{1.0\textwidth}{!}{
%     \includegraphics[]{fig/appendix-example.pdf}
% }
% \caption{Examples of low-entropy and high-entropy example.}
% \end{figure*}
    \noindent\begin{minipage}{0.99\textwidth}
        \centering
        \includegraphics[width=0.99\linewidth]{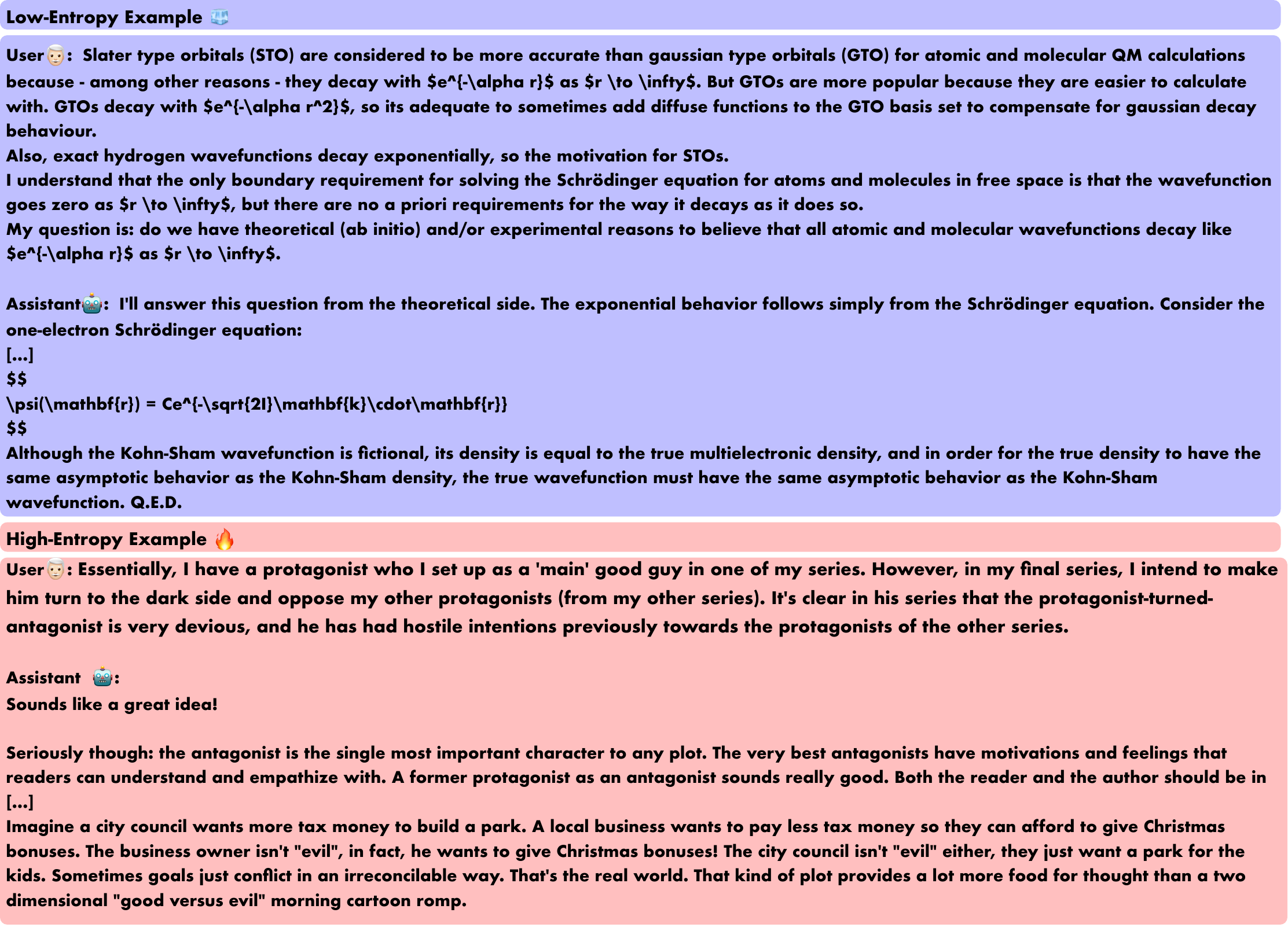} % Replace with your figure
        \captionof{figure}{Examples of low-entropy and high-entropy example.}
        \label{fig:appendix-figure1}
    \end{minipage}

\FloatBarrier
\section{Uncertainty Statistics}\label{appendix:uncertainty-stat}
% \begin{figure*}[b]
% \centering
% \resizebox{.9\textwidth}{!}{
%     \begin{subfigure}{.45\textwidth}
%     \includegraphics[width=\linewidth]{fig/appendix-score-deita.pdf}
%     \end{subfigure}
%     \begin{subfigure}{.45\textwidth}
%     \includegraphics[width=\linewidth]{fig/appendix-score-lima.pdf}
%     \end{subfigure}
% }
% \caption{Uncertainty estimation by GPT-4. \textsc{Left}: Uncertainty scores of Deita-6k alignment dataset. \textsc{Right}: Uncertainty scores of LIMA dataset.}
% \label{fig:uncertainty-stat}
% \end{figure*}

\noindent\begin{minipage}{1.\textwidth}
\includegraphics[width=0.5\linewidth]{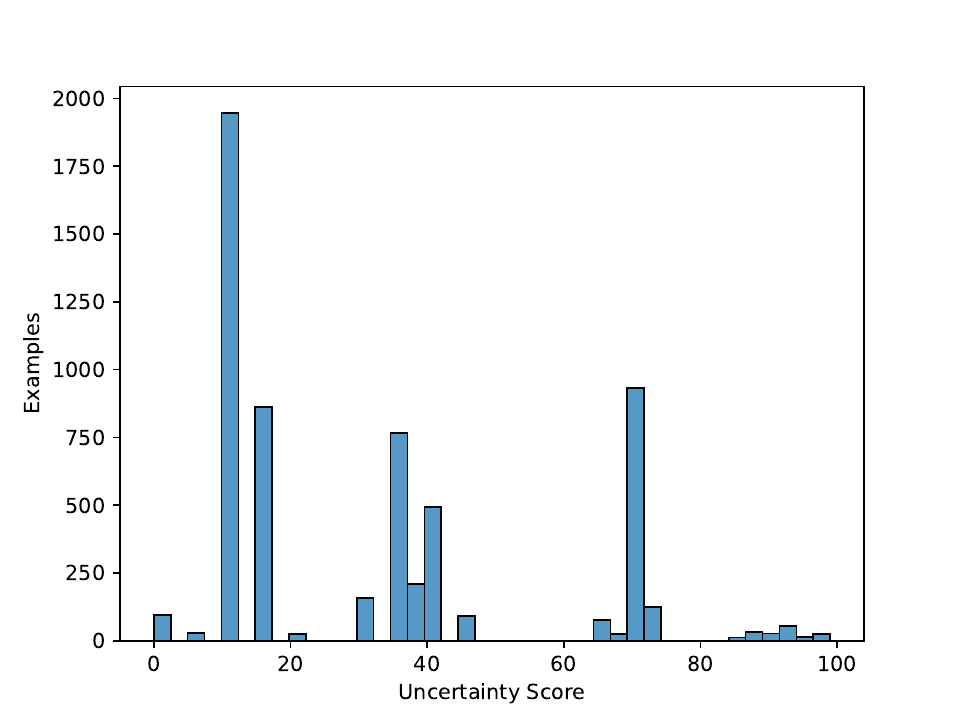}
\includegraphics[width=0.5\linewidth]{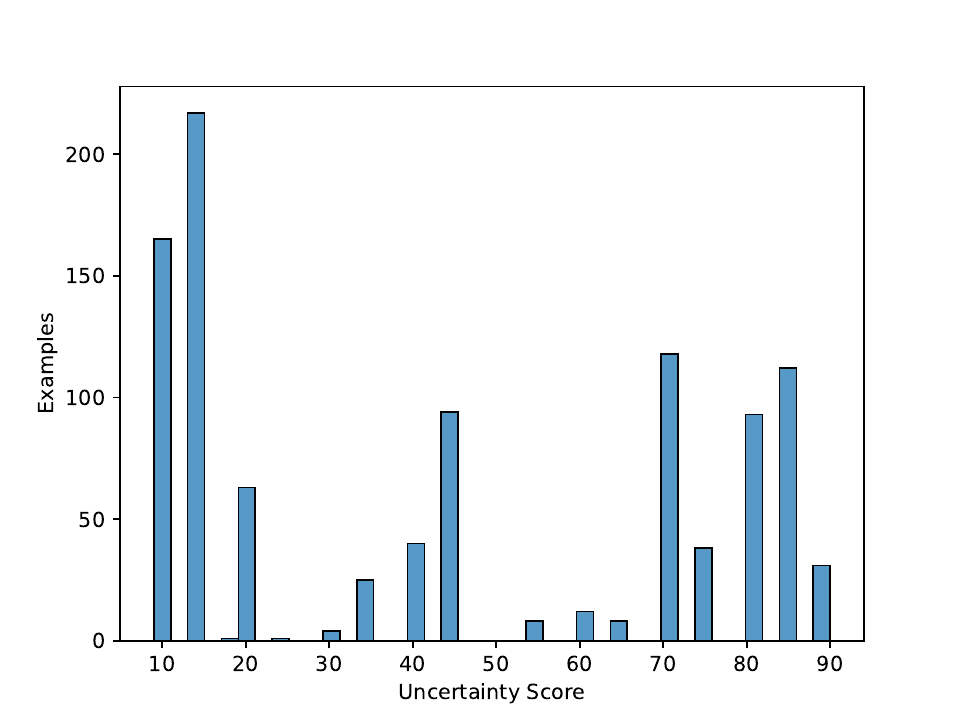}
\captionof{figure}{Uncertainty estimation by GPT-4. \textsc{Left}: Uncertainty scores of Deita-6k alignment dataset. \textsc{Right}: Uncertainty scores of LIMA dataset.}
\end{minipage}
\clearpage

\FloatBarrier
\section{Prompt Examples}\label{appedix:prompt}
% \begin{figure*}[t]
% \centering
% \resizebox{1.0\textwidth}{!}{
%     \includegraphics[]{fig/appendix-sec-prompt.pdf}
% }
% \caption{Prompt examples incorporating OpenAI GPT-4.}
% \label{fig:prompt}
% \end{figure*}
\noindent\begin{minipage}{\textwidth}
        \centering
        \includegraphics[width=0.85\linewidth]{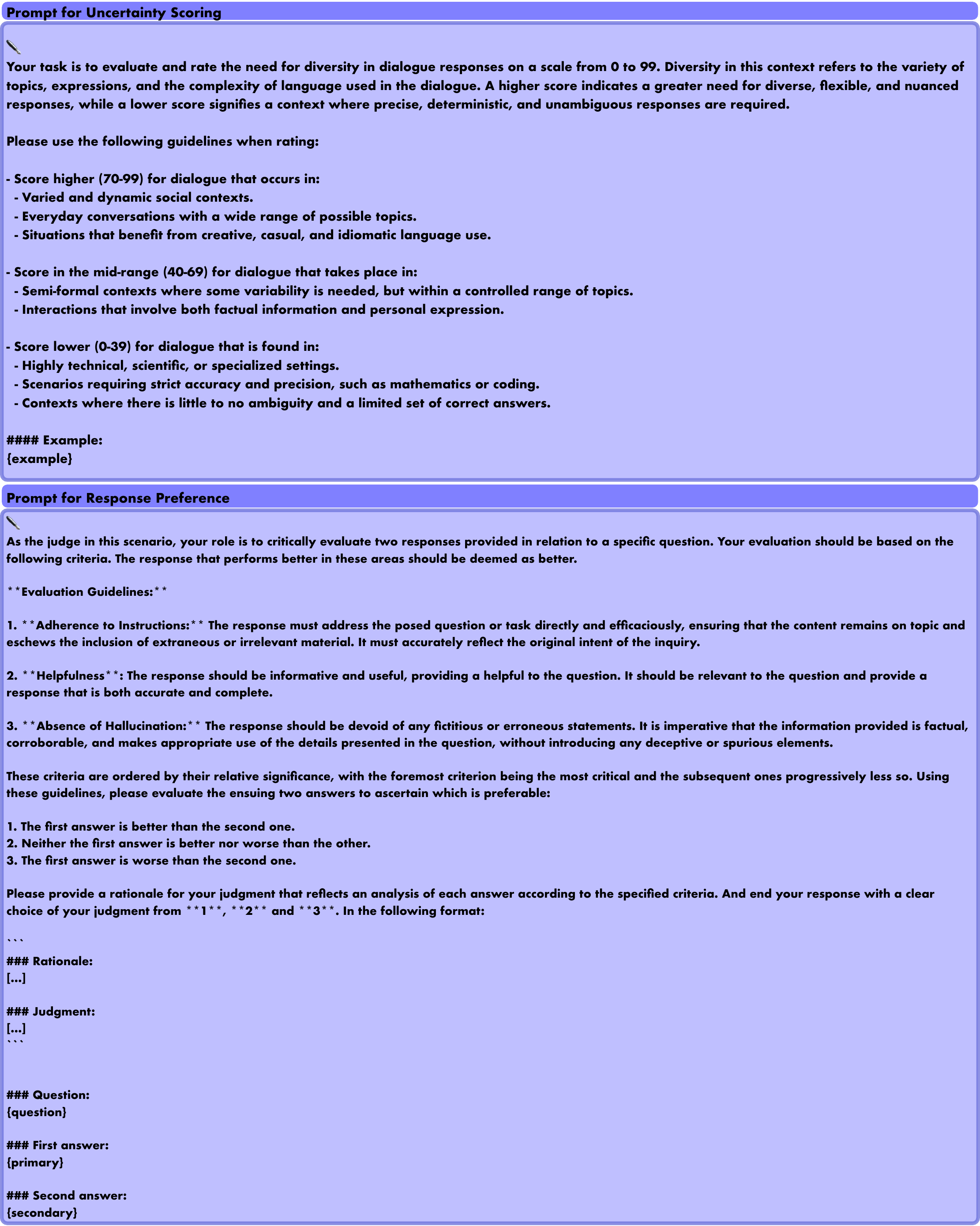} % Replace with your figure
        \captionof{figure}{Examples of low-entropy and high-entropy example.}
        \label{fig:appendix-prompts}
    \end{minipage}

\FloatBarrier
\section{Hyperparameters setting}\label{appendix:hyper}
The foundation models employed are Llama2-7B and Mistral-7B, which are aligned using the LoRA technique if without any other specifications. We set the bias to none, with LoRA parameters $r = 8$ and $\alpha = 16$. Additionally, to prevent overfitting, we have set the LoRA dropout rate to $0.1$.

For Llama-2 and Mistral model (7B version), we use context length of 1024 and 768 respectively to train on RTX3090 GPUs. For Deita-6k dataset we apply $\alpha=0.1$ for UAL-aligned models and larger $\alpha$ (i.e. $0.3$) for LIMA dataset. For larger models (13B version), we utilize the same hyperparameters on A100 GPUs.
\clearpage

\end{document}